\documentclass[11pt,a4paper]{article}
\usepackage[hyperref]{emnlp2020}
\usepackage{times}
\usepackage{latexsym}

\usepackage{microtype}
\usepackage[utf8]{inputenc}
\usepackage{dirtytalk}
\usepackage{natbib}
\usepackage{amsmath}
\usepackage{amsfonts}
\usepackage{bm}
\usepackage{adjustbox}

\usepackage{xcolor}
\usepackage{xspace}
\usepackage{comment}
\usepackage{graphicx}
\usepackage{url}
\usepackage{array}
\usepackage{multirow}
\usepackage{booktabs}
\usepackage{caption}
\usepackage{subcaption}
\usepackage{etoolbox}

\aclfinalcopy 

\newcommand{\joy}[1]{{\color{blue}{\bf #1 -joy}}}
\newcommand{\cy}[1]{{\color{orange}{\bf #1 -cy}}}

\newcommand{\kenneth}[1]{{\color{blue}{\bf #1 -Kenneth}}}

\newcommand{\fitb}{context modeling \xspace}
\newcommand{\ent}{entailment modeling \xspace}

\newcommand{\entshort}{EMLA\xspace}
\newcommand{\fitbshort}{CMLA\xspace}

\title{Assessing the Helpfulness of Learning Materials \\with Inference-Based Learner-Like Agent}

\author{Yun-Hsuan Jen$^1$, Chieh-Yang Huang$^2$, Mei-Hua Chen$^3$, \\
\textbf{Ting-Hao (Kenneth) Huang$^2$, and Lun-Wei Ku$^1$} \\
  $^1$Academia Sinica, Taipei, Taiwan. \\ \texttt{yhjen2@gmail.com}, \texttt{lwku@iis.sinica.edu.tw}\\
  $^2$Pennsylvania State University, University Park, PA, USA. \\ \texttt{\{chiehyang,txh710\}@psu.edu}\\
  $^3$Tunghai University, Taichung, Taiwan. \texttt{mhchen@thu.edu.tw}
}

\date{}

\begin{document}
\maketitle
\begin{abstract}
Many English-as-a-second language learners have trouble using near-synonym words ({\em e.g.,} small vs. little; briefly vs. shortly)
correctly, and often look for example sentences to learn how two nearly synonymous terms differ.
Prior work uses hand-crafted scores to recommend sentences but has difficulty in adopting such scores to all the near-synonyms as near-synonyms differ in various ways.
We notice that the helpfulness of the learning material would reflect on the learners' performance.
Thus, we propose the inference-based learner-like agent to mimic learner behavior and identify good learning materials by examining the agent's performance.
To enable the agent to behave like a learner, we leverage entailment modeling's capability of inferring answers from the provided materials.
Experimental results show that the proposed agent is equipped with good learner-like behavior to achieve the best performance in both fill-in-the-blank (FITB) and good example sentence selection tasks. We further conduct a classroom user study with college ESL learners.
The results of the user study show that the proposed agent can find out example sentences that help students learn more easily and efficiently.
Compared to other models, the proposed agent improves the score of more than 17\% of students after learning.

\end{abstract}

\section{Introduction}
\label{sec:intro}

Many English-as-a-second-language (ESL) learners have trouble using near-synonyms correctly~\cite{10.1093/applin/amu022,liu2013using}.
``Near-synonym'' refers to a word whose meaning is similar but not identical to that of another word, for instance, \textit{establish} and \textit{construct}.
An experience common to many ESL learners is looking for example sentences to learn how two nearly synonymous words differ~\cite{liu2013using,doi:10.1111/j.1540-4781.2009.00828.x}.
To facilitate the learner's learning process, our focus is on finding example sentences to clarify English near-synonyms.

\begin{figure}[t]
    \centering
    \includegraphics[width=0.9\linewidth]{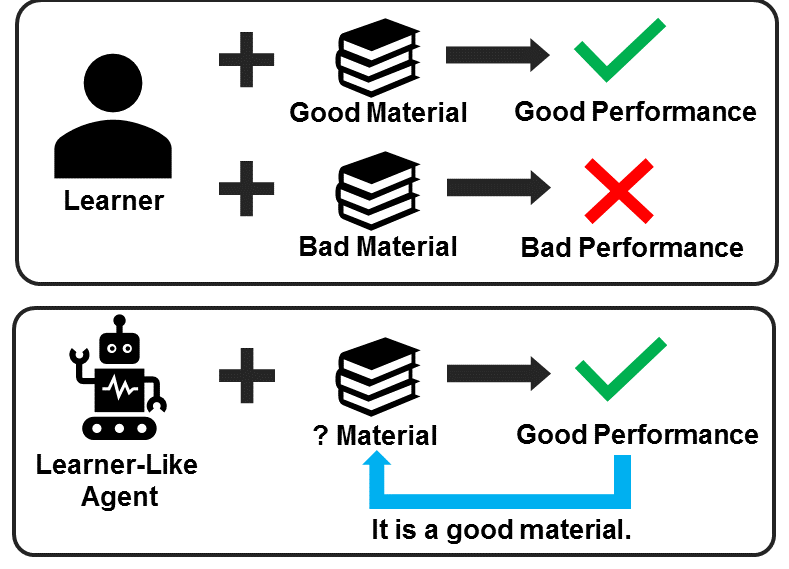}
    \caption{The Learner-Like Agent mimics learners' behavior of performing well when learning from good material and vice versa. We utilize such a behavior to find out helpful learning materials.}
    \vspace{-4mm}
    \label{fig:lela}
\end{figure}

In previous work, researchers develop linguistic search engines, such as Linggle~\cite{boisson2013linggle} and Netspeak\footnote{Netspeak: www.netspeak.org}, to allow users to query English words in terms of n-gram frequency.
However, these tools can only help people investigate the difference, where learners are required to make assumptions toward the subtlety and verify them with the tools, but can not tell the difference proactively.
Other work attempts to automatically retrieve example sentences for dictionary entries ~\cite{kilgarriff2008gdex}; however, finding clarifying examples for near-synonyms is not the goal of such work. 
In a rare exception, \citet{huang2017towards} retrieve useful examples for near-synonyms by defining a clarification score for a given English sentence and using it to recommend sentences.
However, the sentence selection process depends on handcrafted scoring functions that are unlikely to work well for all near-synonym sets.
For example, the difference between \textit{refuse} and \textit{reject} is their grammatical usages where we would use ``refuse to verb'' but not ``reject to verb'';
such a rule, yet, is not applicable for \textit{delay} and \textit{postpone} as they differ in sentiment where \textit{delay} expresses more negative feeling.
Though \citet{huang2017towards} propose two different models to handle these two cases respectively, there is no clear way to automatically detect which model we should use for an arbitrary near-synonym set.

In the search for a better solution, we noted that ESL learners learn better with useful learning materials---as evidenced by their exam scores---whereas bad materials cause confusion.
Such behavior can be used to assess the \textit{usefulness} of example sentences as shown in Figure~\ref{fig:lela}.
Therefore, we propose a \textbf{Learner-Like Agent} which mimics human
learning behavior to enable the ability to select good example sentences.
This task concerns the ability to answer questions according to the example sentences for learning.
As such, we transform this research problem to an entailment problem, 
where the model needs to decide whether the provided example sentence can entail the question or not.
Moreover, to encourage learner-like behavior, we propose perturbing instances for model training by swapping the target confusing word to its near-synonyms.
We conduct a lexical choice experiment to show that the proposed \ent can distinguish the difference of near-synonyms.
A behavior check experiment is used to illustrate that perturbing instances do encourage learner-like behavior, that is inferring answers from the provided materials.
In addition, we conduct a sentence selection experiment to show that such learner-like behavior can be used for identifying helpfulness materials.
Last, we conduct a user study to analyze near-synonym learning effectiveness when deploying the proposed agent on students.

Our contributions are three-fold.
We {\em (i)} propose a learner-like agent which perturbs instances to effectively model learner behavior,
{\em (ii)} use inference-based \ent instead of \fitb to discern
nuances between near-synonyms, and
{\em (iii)} construct the first dataset of helpful example sentences for ESL learners.\footnote{Dataset and code are available here: \url{ https://github.com/joyyyjen/Inference-Based-Learner-Like-Agent}}

\section{Related Works}
\label{sec:related}
This task is related to ({\em i}) learning material generation, ({\em ii}) near-synonyms disambiguation, and ({\em iii}) natural language inference.

\textbf{Learning Material Generation.}
Collecting learning material is one of the hardest tasks for both teachers and students.
Researchers have long been looking for methods to generate high-quality learning material automatically.
\citet{Sumita:2005:MNS:1609829.1609839,sakaguchi-etal-2013-discriminative} proposed approaches to generate fill-in-the-blank questions to evaluate students language proficiency automatically.  
\citet{lin2007automatic,susanti2018automatic,7873236} worked on generating good distractors for multiple-choice questions.
However, there are only a few tasks working on automatic example sentence collection and generation.
\citet{kilgarriff2008gdex,didakowski2012automatic} proposed a set of criteria for a good example sentences and \citet{tolmachev-kurohashi-2017-automatic} used sentence similarity and quality as features to extract high-quality examples.
These tasks only focused on the quality of a single example sentence, whereas our goal in this paper is to generate an example sentence set that clarifies near-synonyms.
The only existing work is from \citet{huang2017towards}, who designed the fitness score and relative closeness score to represent the sentence's ability to clarify near-synonyms.
Our work enables the models to learn the concept of ``usefulness'' directly from data to reduce the possible issues of the human-crafted scoring function. 

\textbf{Near-synonyms Disambiguation.}
Unlike the language modeling task that aims at predicting the next word given the context, near-synonyms disambiguation focuses on differentiating the subtlety of the near-synonyms. 
Edmonds~\shortcite{edmonds1997choosing} first introduced a lexical co-occurrence network with second-order co-occurrence for near-synonym disambiguation. 
Edmonds also suggested a fill-in-the-blank (FITB) task, providing a benchmark for evaluating lexical choice performance on near-synonyms. 
\citet{islam2010near} used the Google 5-gram dataset to distinguish near-synonyms using language modeling techniques.
\citet{wang2010near} encoded words into vectors in latent semantic space and applied a machine learning model to learn the difference.
\citet{huang2017towards} applied BiLSTM and GMM models to learn the subtle context distribution. 
Recently, BERT \cite{devlin2018bert} brought a big success in nearly all the Natural Language Processing tasks.
Though BERT is not designed to differentiate near-synonyms, its powerful learning capability could be used to understand the subtlety lies in the near-synonyms.
In this paper, our models are all designed on top of the pre-trained BERT model.

\textbf{Natural Language Inference.}
Our proposed model directly learns the difference and sentence quality by imitating the human reactions of learning material and behavior of learning from example sentences.
The idea of learning from example is similar to natural language inference (NLI) task and recognizing question entailment (RQE) task. 
There are various NLI dataset varied in size, construction, genre, labels classes ~\cite{bowman2015largE, Williams_2018, Khot2018SciTaiLAT,lai-etal-2017-natural}.
In the NLI task, each instance consists of two natural language text: a premise, a hypothesis, and a label indicating the relationship whether a premise entails the hypothesis. 
RQE, on the other hand, identifies entailment between two questions in the context of question answering. 
\citet{abacha2016recognizing} used the definition of question entailment: ``a question \textit{A} entails a question \textit{B} if every answer to \textit{B} is also a complete or partial answer to \textit{A}.''
Though NLI and RQE research has acquired lots of success, to the best of our knowledge, we are the first to attempt using these two tasks on language learning problems.

\citet{poliak-etal-2018-hypothesis}'s recast version of the definite pronoun resolution (DPR) task inspired us to build learner-like agents with \ent. 
In the original DPR problem, sentences contain two entities and one pronoun, and the mission is to link the pronoun to its referent
\cite{rahman-ng-2012-resolving}. In the recast version, the premises are the original sentences, and the hypothesis is the same sentence with the pronoun replaced with its correct (entailed) and incorrect (not-entailed) reference.
We believe our proposed \ent{} can help the model to understand the relationship between the given example sentence and question for the target near-synonym. 
Thus \ent{} enables the learner-like agent to mimic human behavior through inference.

\section{Method}
\label{sec:ENT}
\begin{table*}[t]
    \centering \small
    \scalebox{0.74}{
        \addtolength{\tabcolsep}{-0.07cm}
        \begin{tabular}{cc|m{8cm}|m{7cm}|p{2cm}}
            \toprule \hline
            \textbf{Model Type} & \textbf{Case} & \multicolumn{1}{c|}{\textbf{Example Sentence}} & \multicolumn{1}{c|}{\textbf{Question}} & \multicolumn{1}{c}{\textbf{Label}} \\ \hline
            \multirow{5}{*}[-2.3em]{\entshort}& (\ref{eqn:eq2}) & After founding the Institute he had \textbf{[little]} time for composing, and appears to have concentrated exclusively on teaching. & When she finds out the truth, she makes a fateful decision to make the most of the \textbf{[little]} time they have together. & \multicolumn{1}{c}{\{$\bm{entail}, \neg \mathit{entail}$\}} \\ \cline{2-5}
            & (\ref{eqn:eq3}) & After founding the Institute he had \textbf{[little]} time for composing, and appears to have concentrated exclusively on teaching. & This may be an incorporated town or city, a subentity of a large city or an unincorporated census-designated place, or a \textbf{[small]} unincorporated community. & \multicolumn{1}{c}{$\{entail, \bm{\neg entail}$\}} \\ \cline{2-5}
            & (\ref{eqn:eq4}) & After founding the Institute he had \textbf{[little]} time for composing, and appears to have concentrated exclusively on teaching. & When she finds out the truth, she makes a fateful decision to make the most of the \textbf{[small]} time they have together. & \multicolumn{1}{c}{$\{entail, \bm{\neg entail}$\}} \\ \cline{2-5}
            & (\ref{eqn:eq5}) & After founding the Institute he had \textbf{[little]} time for composing, and appears to have concentrated exclusively on teaching. & This may be an incorporated town or city, a subentity of a large city or an unincorporated census-designated place, or a \textbf{[little]} unincorporated community. & \multicolumn{1}{c}{$\{entail, \bm{\neg entail}$\}} \\ \cline{2-5}
            & (\ref{eqn:eq8}) & After founding the Institute he had \textbf{[small]} time for composing, and appears to have concentrated exclusively on teaching. & When she finds out the truth, she makes a fateful decision to make the most of the \textbf{[small]} time they have together. & \multicolumn{1}{c}{\{$\bm{entail}, \neg \mathit{entail}$\}} \\ \hline \hline
            
            \multirow{2}{*}[-1em]{\fitbshort}& (\ref{eq:cxt1}) & After founding the Institute he had \textbf{[little]} time for composing, and appears to have concentrated exclusively on teaching. It makes me feel \textbf{[small]} when you keep things from me. & When she finds out the truth, she makes a fateful decision to make the most of the \textbf{\textsc{[MASK]}} time they have together. & \multicolumn{1}{c}{\{$\bm{little}, small$\}} \\ \cline{2-5}
            & (\ref{eq:ctx3}) & It makes me feel \textbf{[little]} when you keep things from me. After founding the Institute he had \textbf{[small]} time for composing, and appears to have concentrated exclusively on teaching. & When she finds out the truth, she makes a fateful decision to make the most of the \textbf{\textsc{[MASK]}} time they have together. & \multicolumn{1}{c}{\{$little, \bm{small}$\}}  \\ \hline \hline
            
            \multicolumn{2}{p{2.4cm}}{Inappropriate Example for \entshort} & After founding the Institute he had \textbf{[small]} time for composing, and appears to have concentrated exclusively on teaching. & When she finds out the truth, she makes a fateful decision to make the most of the \textbf{[little]} time they have together. & \multicolumn{1}{c}{$\{entail, \bm{\neg entail}$\}} \\ \hline
            \multicolumn{2}{p{2.4cm}}{Inappropriate Example for \fitbshort} & It makes me feel \textbf{[little]} when you keep things from me. After founding the Institute he had \textbf{[small]} time for composing, and appears to have concentrated exclusively on teaching. & When she finds out the truth, she makes a fateful decision to make the most of the \textbf{\textsc{[MASK]}} time they have together. & \multicolumn{1}{c}{\{$\bm{little}, small$\}} \\ \hline \bottomrule
        \end{tabular}
        \addtolength{\tabcolsep}{0.07cm}
    }
    \caption{Training instances for learner-like agents. The instances are associated with the corresponding equations. Case (\ref{eqn:eq8}) and (\ref{eq:ctx3}) are the perturbed instances. The inappropriate examples are used in section~\ref{sec:exp} for behavior check.}
    \vspace{-4mm}
    \label{tab:agent_example}
\end{table*}



In this paper, we use \emph{learner-like agent} to refer to a model that answers questions given examples. 
The goal of the learner-like agent is to answer fill-in-the-blank questions on near-synonyms selection.
However, instead of answering the question from the agent's prior knowledge, the agent needs to answer the question using the information from the given examples.
That is, if the given examples provide incorrect information, the agent should then come up with the wrong answer.
This process is to simulate the learner behavior illustrated in the Figure~\ref{fig:lela}.
Since the model is required to infer the answer, we further formulate it as an entailment modeling problem to enable model's capability of inference.
In this section, we ({\em i}) define the proposed learner-like agent, ({\em ii}) describe how to formulate it as an entailment modeling problem, and ({\em iii}) introduce the perturbed instances to further enhance the agent's learner behavior.




\subsection{Learner-Like Agent}
\label{subsec:learner_like_agent}

The overall structure of a learner-like agent is as follows: given six example sentences $\mathbb{E}$ (3 sentences for each word) and a fill-in-the blank question $Q$ as an input instance, the model is to answer the question based on the example hints. 
We adopt BERT~\cite{devlin2018bert} to fine-tune the task-specific layer of the proposed learner-like agent using our training data, equipping the learner-like agent with the ability to discern differences between near-synonyms.
The input of our model contains the following:
\begin{itemize}
\item A question $Q^{w_i} = [q_1, q_2, .., q_n]$, where $n$ is the length of
the sentence and contains a word~$w_i$ from the near-synonym pair, where $i \in
\{1,2\}$ denotes word~1 or word~2;
\vspace{-2mm} 
\item Example sentences set $\mathbb{E} =[E^{w_1}_{1}, ..., E^{w_1}_{3},
E^{w_2}_{4}, ..., E^{w_2}_{6}]$, where $E^{w_i}$ denotes a  sentence containing
$w_i$;  \vspace{-2mm}
\item A \textsc{[cls]} token for the classification position, and several
\textsc{[sep]} tokens used to label the boundary of the question and the
example sentences, following the BERT settings.
\end{itemize}
The output will is the correct word for the input question, namely, $w_1$ or $w_2$.

We specifically define $E[w_j]^i$ where $i, j \in {1,2}$ to be the context of $w_i$. 
The example sentence of case (\ref{eqn:eq2}) in Table~\ref{tab:agent_example} shows a case of $E[w_1]^1$ where the \textbf{target word} $w_1$ is \textit{little} and the rest of the sentence is called \textbf{context} $E[\_]^1$.
When we change \textit{little} to \textit{small} to create case (\ref{eqn:eq8}), it is described as $E[w_2]^1$ meaning an example sentence where $w_2$ fills the position of $w_1$ in sentence~$E^{w_1}$.
This notation also applies to the question input $Q[w_j]^i$.





\subsection{Inference-based Entailment Modeling}

We apply NLI and RQE tasks in the learner-like agent question design. The goal
of the \textbf{E}ntailment \textbf{M}odeling \textbf{L}earner-like \textbf{A}gent (\textbf{\entshort}) is to answer entailment questions given example sentences. 
We transform the original fill-in-the-blank question into an entailment question where the \entshort answers whether the given example sentence $E$ entails the question sentence $Q$. 
If the word usage in the question sentence matches the word usage in the example sentence,
the \entshort answers
$\mathit{entail}$, or $\neg \mathit{entail}$ otherwise. 

The \entshort $M_e$ is described as
\begin{equation}
     M_e(E^{i}_k, Q^{j}) = \mathit{ans},
\end{equation}
where $\mathit{ans}$---either $\mathit{entail}$ or $\neg \mathit{entail}$---is the prediction of
the inference relationship of one of the six example sentences~$E^{i}_k$, where $k \in \{1, 2, ..6\}$, and  $Q^{j}$.
To fill all the context possibilities of $Q[\_]^j$ for the same word in
$E^{w_i}$, an example has the following four cases:
\begin{align}
    &M_e(E[w_1]^1,Q[w_1]^1) = \mathit{entail} \label{eqn:eq2} \\
    &M_e(E[w_1]^1,Q[w_2]^2) = \neg \mathit{entail} \label{eqn:eq3} \\
    &M_e(E[w_1]^1,Q[w_2]^1) = \neg \mathit{entail} \label{eqn:eq4} \\
    &M_e(E[w_1]^1,Q[w_1]^2) = \neg \mathit{entail}. \label{eqn:eq5}
\end{align}

From the input and output of the instances (equations~\ref{eqn:eq2} to
\ref{eqn:eq5}), we see that the target word and its context in $Q^j$ for
all cases except for equation~\ref{eqn:eq2} do not follow the example word usage.
The examples of the instances are shown in Table~\ref{tab:agent_example}.
Equation \ref{eqn:eq3} and equation \ref{eqn:eq4} tell us that an example sentence 
of $w_1$ does not provide any information for the model to infer anything about 
$w_2$ so both of them result in \textit{not entail}.
The question of equation \ref{eqn:eq5} is incorrect, as shown in the Table 1 case (\ref{eqn:eq5}), 
so it would also lead to \textit{not entail}.



After training the \entshort to understand the relation between
example and question, we can convert its prediction \{$\mathit{entail}$, $\neg
\mathit{entail}$\} back into the fill-in-the-blank task by looking into the model
predictions.
Given the probability of \{$\mathit{entail}$, $\neg \mathit{entail}$\}, we 
know which term in the near-synonym pair is more appropriate in the context of 
$\{Q[\_]^1, Q[\_]^2\}$. If the question context and the example context match,
then a word with a higher $\mathit{entail}$ probability is the answer. If
they do not match, that with the higher $\neg \mathit{entail}$ probability is the
answer. 

\subsection{Perturbed Instances}

\label{subsec:perturbing_example}
To encourage learner-like behavior, i.e., good examples lead to the correct
answer, and vice versa, we propose introducing automatically generated perturbed
instances to the training process.

A close look at the input and output of the instances 
(equations~\ref{eqn:eq2} to \ref{eqn:eq5}) shows that they consider only correct examples
and their corresponding labels. We postulate that wrong word usage yields inappropriate
examples; thus we perturb instances by swapping the current
confusing word to its  near-synonym as 
\begin{equation}\vspace{-2mm}
    M_e(E[\neg w_i]^{i}_k, Q^{w_j}) = \mathit{\neg ans}
\end{equation}
where $\mathit{\neg ans}$ is 
  $\{entail,\neg entail\} - \mathit{ans}$    
and
$E[\neg w_i]^{w_i}_k$ is the example sentence in which the contexts in 
$w_1$ and $w_2$ are swapped. 
The corresponding perturbed instances from equations~\ref{eqn:eq2}
to \ref{eqn:eq5} thus become
\begin{align}
    &M_e(E[w_2]^1,Q[w_1]^1) = \neg \mathit{entail} \label{eqn:eq6}\\
    &M_e(E[w_2]^1,Q[w_2]^2) = \neg \mathit{entail} \label{eqn:eq7}\\
    &M_e(E[w_2]^1,Q[w_2]^1) = \mathit{entail} \label{eqn:eq8}\\
    &M_e(E[w_2]^1,Q[w_1]^2) = \neg \mathit{entail}, \label{eqn:eq9}
\end{align}
respectively, in which $w_2$'s context becomes $E[\_]^1$.
Again, only equation~\ref{eqn:eq8}, where both the context and the word usage match, is \textit{entail}. 
The example instance is shown in Table~\ref{tab:agent_example} case~\ref{eqn:eq8}.

\vspace{-1mm}
\section{Experiments}
\label{sec:exp}
\vspace{-1mm}
We conducted three experiments: \textbf{lexical choice}, \textbf{behavior
check}, and \textbf{sentence selection}. 
The lexical choice task assesses whether the model differentiates confusing
words,
the behavior check measures whether the model responds to the quality of
learning material as learners do, 
and sentence selection evaluates the model's ability to explore useful example
sentences.


\subsection{Lexical Choice}
Lexical choice evaluates the model's ability to differentiate
confusing words.
We adopted the fill-in-the-blank (FITB) task, where the model is asked to choose a word
from a given near-synonym word pair to fill in the blank.

\subsubsection{Baseline}
\label{subsec:FITB}
\expandafter\MakeUppercase\fitb{} is a common practice for near-synonym
disambiguation in which the model learns the context of the target word via
the FITB task. 
For this we use a \textbf{C}ontext \textbf{M}odeling \textbf{L}earner-like \textbf{A}gent (\textbf{\fitbshort}) as the baseline based on 
BERT~\cite{devlin2018bert} as a two-class classifier to predict which of
$w_1$ or $w_2$ is more appropriate given a near-synonym word pair. 
The question for \fitbshort is a sentence whose target word,
i.e., one of the confusing words, is masked; the model is to predict the
masked target word. 


The \fitbshort $M_c$ is then described as
\begin{equation}
    M_c(\mathbb{E}, Q[\textsc{mask}]^{i}) = \mathit{ans},
\end{equation}
where $Q[\textsc{mask}]^{i}$ fills the the position of ${w_i}$ with
\textsc{mask}, and ${\mathit{ans}} \in \{w_1,w_2\}$ is the prediction of 
\textsc{[mask]} in the question, and $\mathbb{E}$ are the six example sentences. 

  $Q[\textsc{mask}]^{i}$ is a question with the context of either ${w_1}$ or ${w_2}$.   
This raises a problem of the model deriving the answer only from $Q^i$,  
\begin{align}
\vspace{-3mm}
   & M_c(\mathbb{E}, Q[\textsc{mask}]^1) = w_1 \label{eq:cxt1}\\
   & M_c(\mathbb{E}, Q[\textsc{mask}]^2) = w_2
\label{eq:cxt2}
\vspace{-3mm}
\end{align}
Equations~\ref{eq:cxt1} and~\ref{eq:cxt2} risk the model to selects $w_i$ given $Q^i$. 
To encourage learner-like behavior, we incorporate perturbed
instances into the training process corresponding to equations~\ref{eq:cxt1}
and \ref{eq:cxt2} as 
\vspace{-3mm}
\begin{align}
    &  M_c(\neg \mathbb{E}, Q[\textsc{mask}]^1) = w_2 \label{eq:ctx3}\\
    &  M_c(\neg \mathbb{E}, Q[\textsc{mask}]^2) = w_1,  \label{en:ctx4}\vspace{-3mm}
\end{align}
, where $\neg \mathbb{E}= [E[\neg w_2]^{2}_1,..,E[\neg w_2]^{2}_3,E[\neg w_1]^{1}_1,..,E[\neg w_1]^{1}_3]$

For \fitb{}, the perturbed instance has the additional benefit that it
forces the model to make inferences based on the given example sentences, as illustrated in Table~\ref{tab:agent_example} case (\ref{eq:ctx3}).




\subsubsection{Dataset and Settings}
We collected a set of near-synonym word pairs from online resources, including
BBC\footnote{\scriptsize{http://www.bbc.co.uk/learningenglish/chinese/features/q-and-a}},
the Oxford Dictionary\footnote{\scriptsize{https://en.oxforddictionaries.com/usage/commonly-confused-words}}, 
and a Wikipedia page about commonly misused English words\footnote{\scriptsize{https://en.wikipedia.org/wiki/Wikipedia:List\_of\_commonly\_\\misused\_English\_words}}.

An expert in ESL education manually selected 30 near-synonym word pairs 
as our experimental material.
We collected our data for both training and testing from Wikipedia on January~20, 
2020. Words in the confusing word pair were usually of a specific part of
speech. This guaranteed that the part of speech of the confusing word in the
sentence pool matched that in target near-synonym word pair. 
To construct a balanced dataset, we randomly selected 5,000 sentences for each
word; 4,000 sentences for each word in a near-synonym word pair were used to
train the learner-like model and 1,000 sentences for testing. 

For comparison, we trained four learner-like agents: \entshort, \fitbshort, \entshort without perturbed
instances, and \fitbshort without perturbed instances. For the best learning effect, we
empirically set the ratio of normal-to-perturbed instances to $2:1$.
The agents were trained using the Adam optimizer with a 30\% warm-up
ratio and a 5e-5 learning rate. The maximum total input sequence length
after tokenization was 256; other settings followed the BERT configuration. 

\subsubsection{Results and Discussion}
\begin{figure}[t]
    \centering
    \includegraphics[width=1.0\columnwidth]{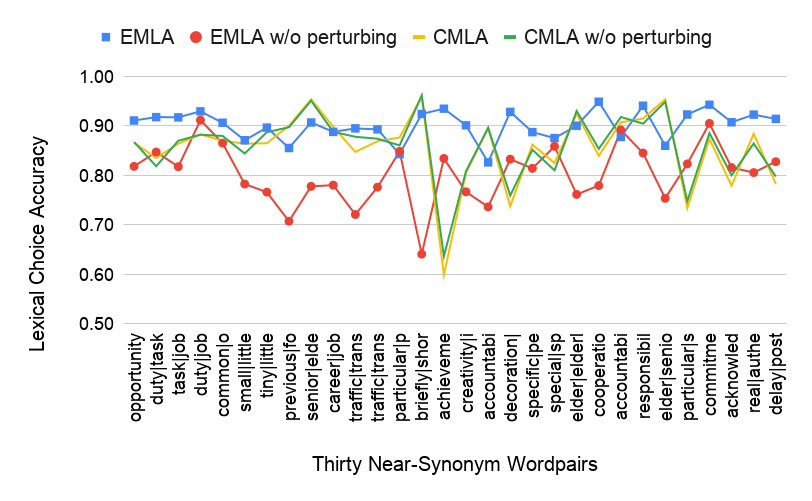}
	 \caption{Visualization of lexical choice performance on near-synonym word pairs. Adding perturbed instances improves \entshort's ability of differentiating confusing words. However, perturbed instances does not cause big difference on \fitbshort.}
    \label{fig:model_performance}
\end{figure}

We compared the \entshort and \fitbshort and
Figure~\ref{fig:model_performance} shows the model performance on 30 word
pairs. The average accuracy of \entshort and \fitbshort is 0.90 and 0.86, while that
excluding perturbing instances is 0.80 and 0.86, respectively. 
On average, \entshort performs the best; when perturbed instances are not
included in the training, its performance for lexical choice drops.  
We expected training with perturbed instances to worsen model
performance in exchange for learner-like behavior. 
However, results show that the perturbed instances enhance the inference ability of \entshort.
Also, \fitbshort models seem to be unaffected by perturbed
instances (yellow vs. green lines); this could be because
\fitbshort tends to memorize the input context 
  instead of making an actual inference,   
which in NLI is recognized as bias~\cite{Chien_2020}. 

\subsection{Behavior Check}
\label{subsec:behave-check}

\begin{table}[t]
\small \center
\begin{tabular}{lcc}
\toprule \hline
                   & \textbf{T-score} & \textbf{P-value} \\ \hline
\textbf{\fitbshort}              & 24.54    & 2.06e-21 \\ 
\textbf{\fitbshort w/o perturbing}     & 0.77     & 0.45     \\ 
\textbf{\entshort}         & 92.12    & 2.43e-37 \\ 
\textbf{\entshort w/o perturbing} & 27.06    & 3.96e-22 \\ \hline \bottomrule
\end{tabular}
\caption{Except for \fitb{} without perturbed instances, all models respond to 
changes in learning material quality}
\label{table:behavior_check_result}
\vspace{-3mm}
\end{table}

The behavior check evaluates whether the agent learns as learners do; that is, a
learner-like agent should perform well on 
FITB questions when the given learning materials are helpful, and should
perform poorly when the materials are not helpful.

In this experiment, all models complete two FITB quizzes.
For the first quiz, authentic sentences are provided as appropriate
learning materials; for the second quiz, inappropriate learning materials are provided.
These materials are considered inappropriate because they are
automatically generated using the authentic sentences but replacing their
target words with near-synonyms for training, resulting in confusion and wrong
word usage, as illustrated in Table~\ref{tab:agent_example} (see the last two ``Inappropriate example'' rows). 
In other words, given inappropriate example sentences, if the model is truly inferring answers from
the examples, the model should select the other choice in the same quiz question.

\begin{figure*}[t]
    \centering
    \begin{subfigure}[b]{.42\linewidth}
    \includegraphics[width=1.0\columnwidth]{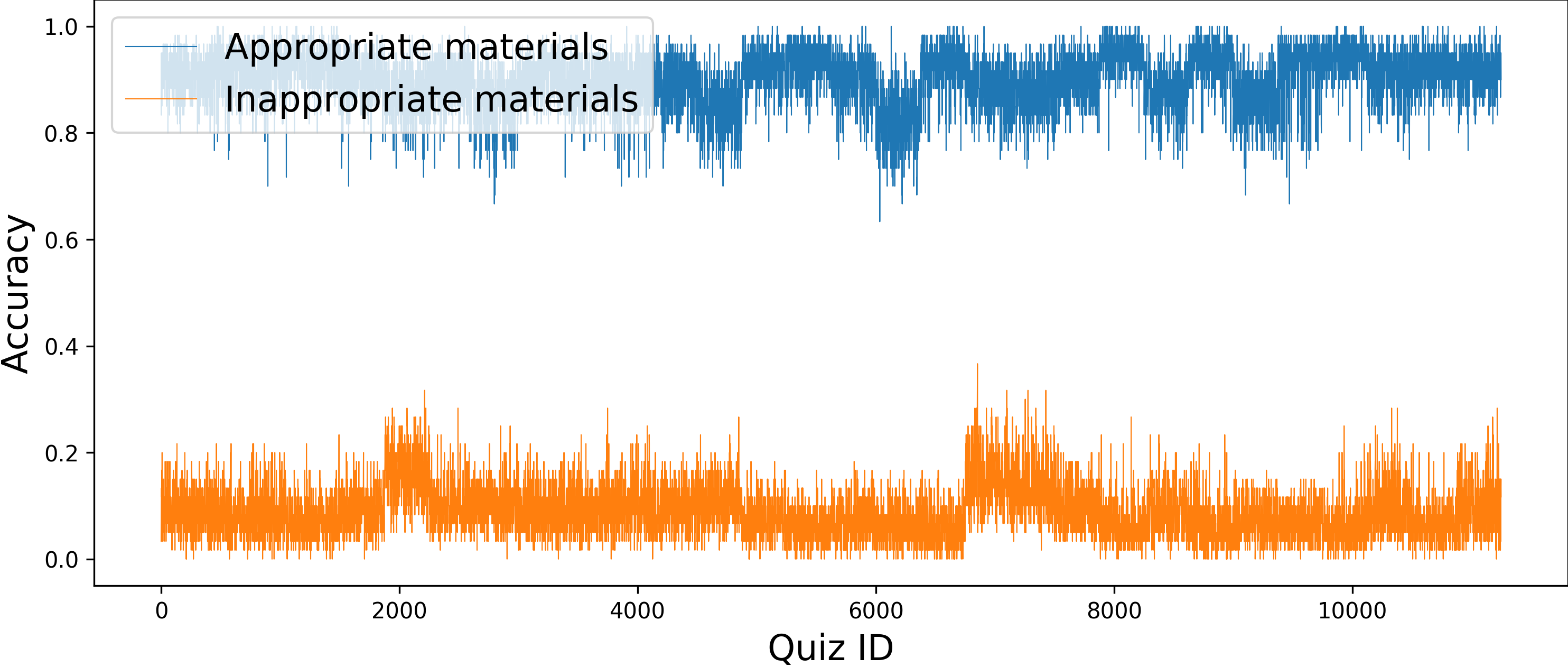}
    \caption{\entshort}
    \end{subfigure}
    \begin{subfigure}[b]{.42\linewidth}
    \includegraphics[width=1.0\columnwidth]{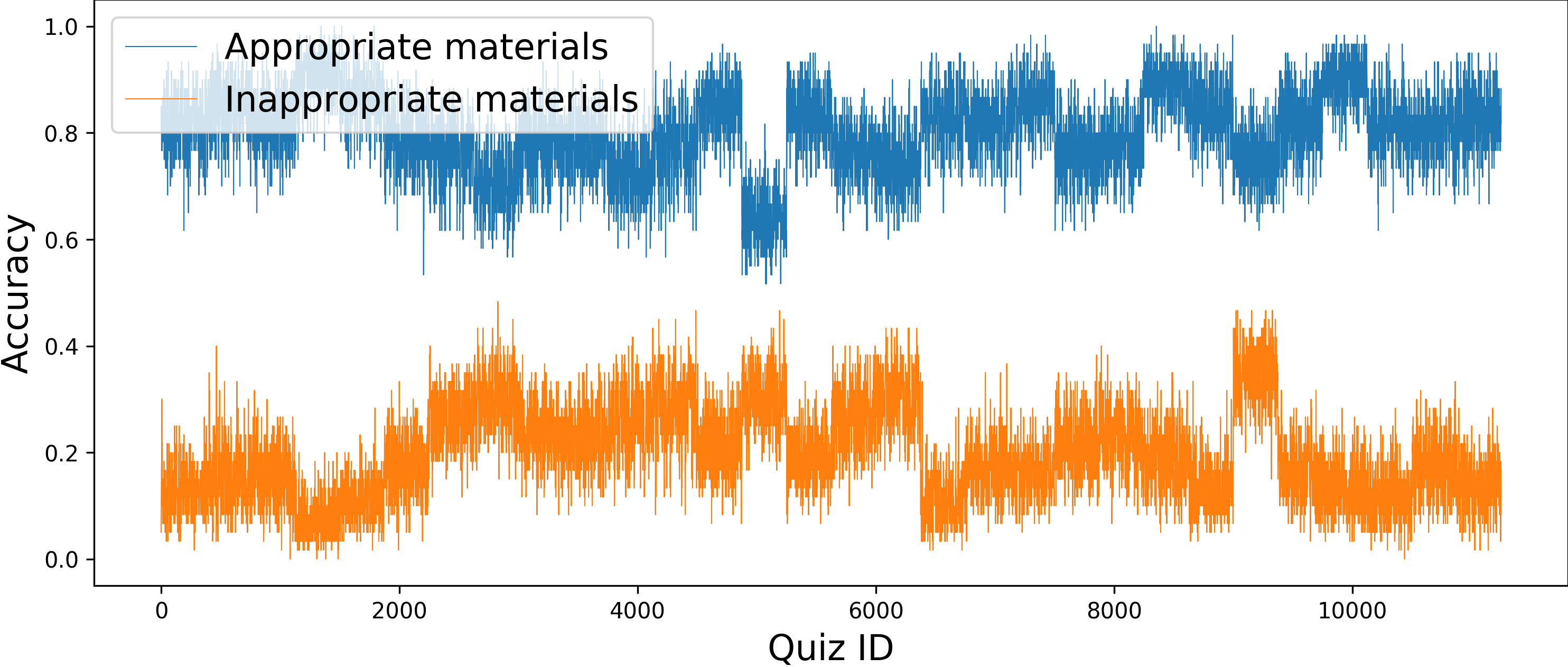}
    \caption{\entshort w/o perturbing}
    \end{subfigure}
    \begin{subfigure}[b]{.42\linewidth}
    \includegraphics[width=1.0\columnwidth]{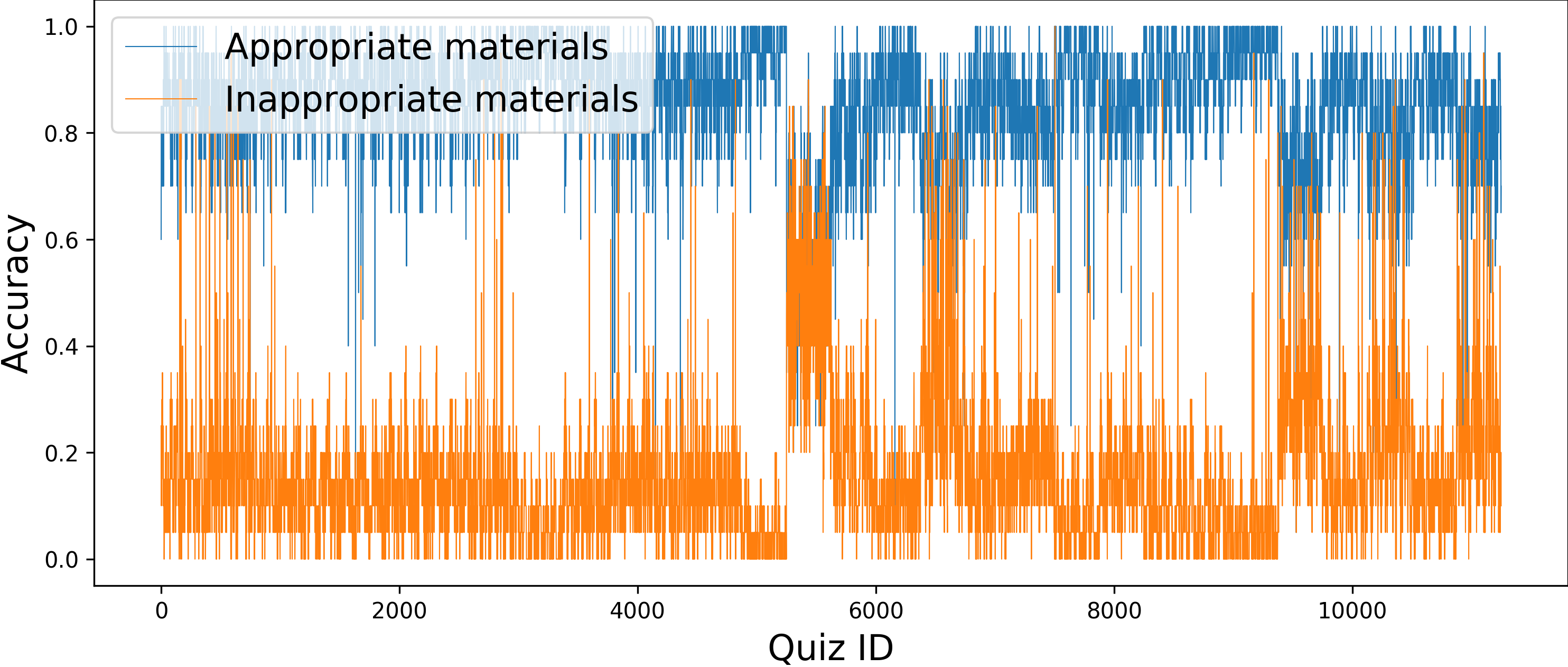}
    \caption{\fitbshort}
    \end{subfigure}
    \begin{subfigure}[b]{.42\linewidth}
    \includegraphics[width=1.0\columnwidth]{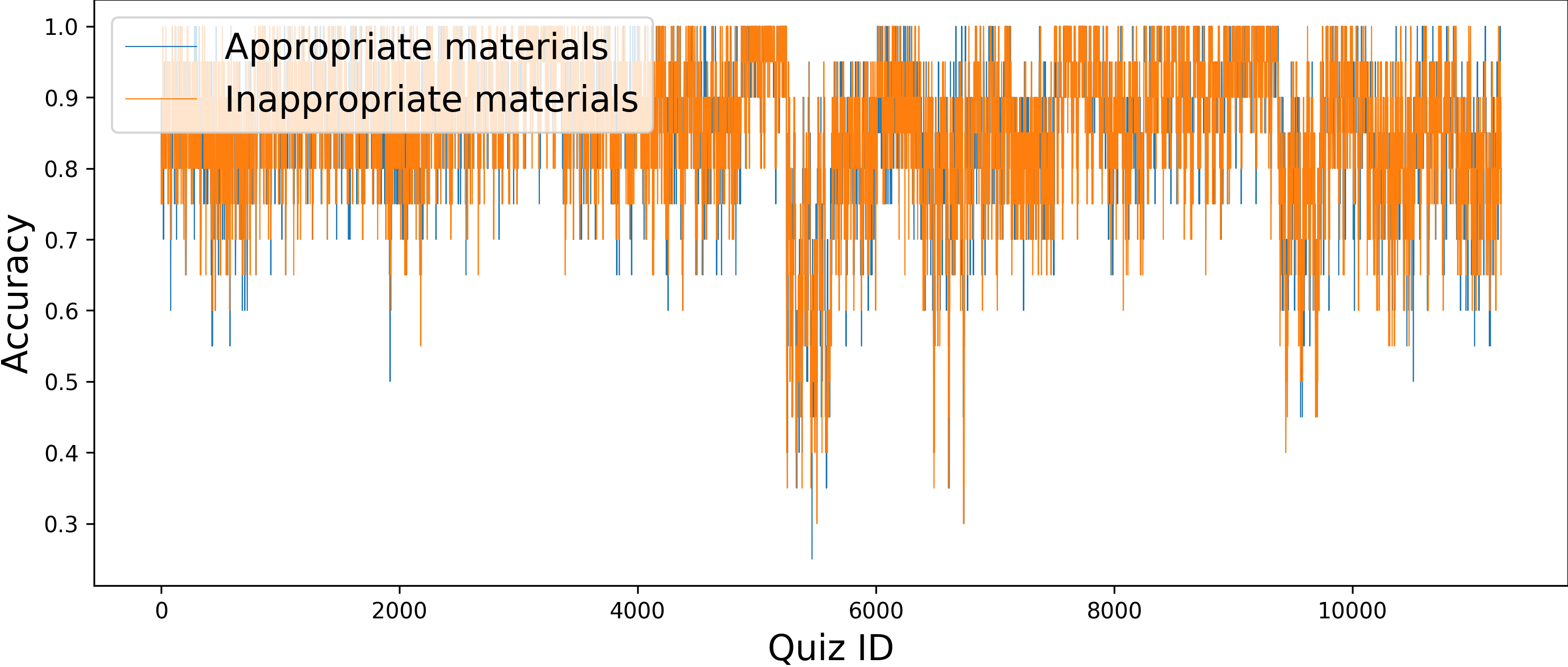}
     \caption{\fitbshort w/o perturbing}
     \end{subfigure}

	 \caption{Behavior check visualization of quiz results given appropriate (blue) or inappropriate (orange)
	 learning materials. 
	 Each quiz was completed 11250 times (375 example sentence sets for each of the 30 
 	 word pairs).}   
    \label{fig:behavior_check}
\vspace{-2mm}
\end{figure*}

\subsubsection{Results and Discussion}


We recorded the accuracy of every question and combined the 30 pairs of near-synonym
wordsets from the same model into one graph. As shown in 
Figure~\ref{fig:behavior_check}, even without perturbed instances, the
learning effect of \entshort corresponds to the learning material quality. In 
contrast, \fitbshort without perturbed instances, as in the lexical
choice task, is no worse when given inappropriate examples.

To determine whether the results of the two fill-in-the-blank quizzes
are significantly different when given appropriate and inappropriate examples, we conducted a
t-test. Table~\ref{table:behavior_check_result} shows that 
learner-like behavior is enabled in \fitbshort with perturbed instances, whereas
\entshort learns like learners even without perturbed instances. This result
conforms to that shown in Figure~\ref{fig:behavior_check}: the quiz results 
for both \entshort models can be clearly distinguished, and adding the
perturbed instances to \entshort slightly magnifies their difference. However, the
\fitbshort still relies on perturbed instances to learn the difference. 

Looking more closely, we present Table~\ref{tab:behavior_result_by_wordpair}, in which
$\Delta$ is the difference in accuracy between two quizzes. The higher $\Delta$ is, the
better the model differentiates confusing words. We measure the
correlation between the lexical choice accuracy and $\Delta$ with the Pearson
correlation coefficient and obtain a value of 0.87, which demonstrates a strong
positive correlation.

\begin{table*}[]
\center
\scalebox{0.75}{
\begin{tabular}{|c|c|c|c|c|c|c|c|c|}
\hline
                              & \textbf{Acc} & \textbf{$\Delta$} &                         & \textbf{Acc} & \textbf{$\Delta$} &                            & \textbf{Acc} & \textbf{$\Delta$} \\ \hline
accountability, responsibility & 0.83   & 0.76  & traffic, transportation  & 0.89   & 0.80  & duty, task                  & 0.92   & 0.84  \\ \hline
particular, peculiar           & 0.84   & 0.73  & tiny, little             & 0.90   & 0.79  & real, authentic             & 0.92   & 0.86  \\ \hline
previous, former               & 0.86   & 0.76  & elder,elderly           & 0.90   & 0.80  & particular, specific        & 0.92   & 0.86  \\ \hline
elder, senior                  & 0.86   & 0.79  & creativity, innovation   & 0.90   & 0.83  & briefly, shortly            & 0.92   & 0.85  \\ \hline
small, little                  & 0.87   & 0.71  & common, ordinary         & 0.91   & 0.82  & decoration, ornament        & 0.93   & 0.87  \\ \hline
special, specific              & 0.88   & 0.74  & senior, elderly          & 0.91   & 0.82  & duty, job                   & 0.93   & 0.86  \\ \hline
accountability, liability      & 0.88   & 0.79  & acknowledge, admit       & 0.91   & 0.81  & achievement, accomplishment & 0.93   & 0.88  \\ \hline
specific, peculiar             & 0.89   & 0.73  & opportunity, possibility & 0.91   & 0.82  & responsibility, liability   & 0.94   & 0.88  \\ \hline
career, job                    & 0.89   & 0.80  & delay, postpone          & 0.91   & 0.82  & commitment, responsibility  & 0.94   & 0.88  \\ \hline
traffic, transport             & 0.89   & 0.80  & task, job                & 0.92   & 0.83  & cooperation, collaboration  & 0.95   & 0.89  \\ \hline
\end{tabular}}
\caption{Mapping of lexical performance and model ability to differentiate
near-synonyms for 30 word pairs using \ent{}. Acc is the lexical choice accuracy on appropriate examples and $\Delta$ is the difference in accuracy between the two quizzes.}
\vspace{-5mm}
\label{tab:behavior_result_by_wordpair}
\end{table*}

\subsection{Sentence Selection}
\label{subsec:sentence-selection}
In the sentence selection experiment, we evaluate the ability of the
learner-like agent to select useful example sentences. Our assumption is
straightforward. We give the agent a set of example
sentences and evaluate its performance on a number of quizzes. If it does
well on many quizzes, the example sentences are deemed helpful for learning confusing
words. 

\subsubsection{Baseline}
We compared agents with an implementation of
\citet{huang2017towards}'s Gaussian mixture model (GMM), which
learns the distribution and semantics of the context. We set the
number of Gaussian mixtures to 10 and trained the GMM with the dataset proposed here.
In the testing phase, we retrieved the top three recommended sentences
for each word in the confusing word pair and compared this to the expert's
choices.
\vspace{-1mm}
\subsubsection{Evaluation Dataset}
To evaluate the sentence selection, we employed an ESL teacher as an expert to
carefully select the three best example sentences out of ten randomly selected,
grammatically, and pragmatically correct examples for each word in all
confusing word pairs. 
 
Specifically, the evaluation dataset had a total of 600 example sentences. For
each near-synonym pair, three sentences for each word were labeled as helpful
example sentences. To select sentences that clearly clarify the semantic
difference between near-synonyms, the ESL expert considered
\textit{suitability}, \textit{informativeness}, \textit{diversity},
\textit{sentence complexity}, and \textit{lexical complexity} during
selection. For \textit{suitability}, the expert considered whether the two
near-synonym words in one confusing word pair were interchangeable in the current
sentence. \textit{Diversity} was considered when constructing the selected pool.
\textit{Suitability} and \textit{diversity} are designed from the \citep{huang2017towards}'s conclusion.
Other criteria are from Kilgarriff's good example sentence~\cite{kilgarriff2008gdex}. 
 
\vspace{-1mm}
\subsubsection{Selection Method}
For the proposed good example sentence set, we selected an example sentence 
combination that helps \entshort or \fitbshort to achieve the highest accuracy 
in the quiz. That is, the example sentence set that leads to the highest learning performance.

One of a total of 14,400 ($C^{10}_3 \times C^{10}_3$) example sentence sets,
including six example sentences, was provided to the models to evaluate their
helpfulness. 
Each example sentence set was used to answer a quiz composed of $k$ questions.
Here, $k$ determines the representativeness and consistency of the
testing result from each quiz. 
We used five independent quizzes to find a reliable $k$ by calculating the
correlation of their testing results. Finally, we empirically set $k$ to 100,
where the lowest correlation among 30 word pairs was 0.24, and the median was
0.67. That is, each quiz contained 100 questions.

When testing example sentence sets, multiple example sentence sets could achieve 
the same highest accuracy for the quiz. We considered them equally good so sentences 
in these sets were all treated as selected. Thus, our method would possibly
suggest more example sentences than the gold labels.





\subsubsection{Results and Discussion}

Table~\ref{table:sentence_selection_result} shows the results of sentence
selection. 
\entshort significantly outperforms \fitbshort and Huang's
GMM in sentence selection. The improvement comes from the increasing recall,
indicating that the proposed learner-like agent manages to find helpful example
sentences for ESL learners. 

\begin{table}[t]
\center
\small
\begin{tabular}{lccc}
\toprule \hline
    & \textbf{Precision} & \textbf{Recall} & \textbf{F1}   \\ \hline
\textbf{\entshort}    & 0.33      & 0.71   & \textbf{0.45} \\ 
\textbf{\fitbshort}      & 0.31      & 0.56   & 0.38 \\ 
\textbf{GMM} & \textbf{0.37}      & 0.34   & 0.35 \\ \hline \bottomrule
\end{tabular}
\caption{Sentence selection results. When compared with the human annotation, \ent achieves the highest F1 of 0.45.}
\label{table:sentence_selection_result}
\vspace{-0.4cm}
\end{table}

\vspace{-1mm}
\section{Learner Study}
\label{sec:learner}
We conducted a user study to see the effect of learning on example sentences selected by \entshort, \fitbshort, and a random baseline.
In this learner study, a total of 29 Chinese-speaking college freshmen majored in English were recruited.
All the participants were aged between 18 and 19. 
A proficiency test~\cite{chen2011factors} was given before the study to identify their English level for further analysis.

\subsection{Experimental Design and Material}
\begin{figure}[t]
    \centering
    \includegraphics[width=0.9\linewidth]{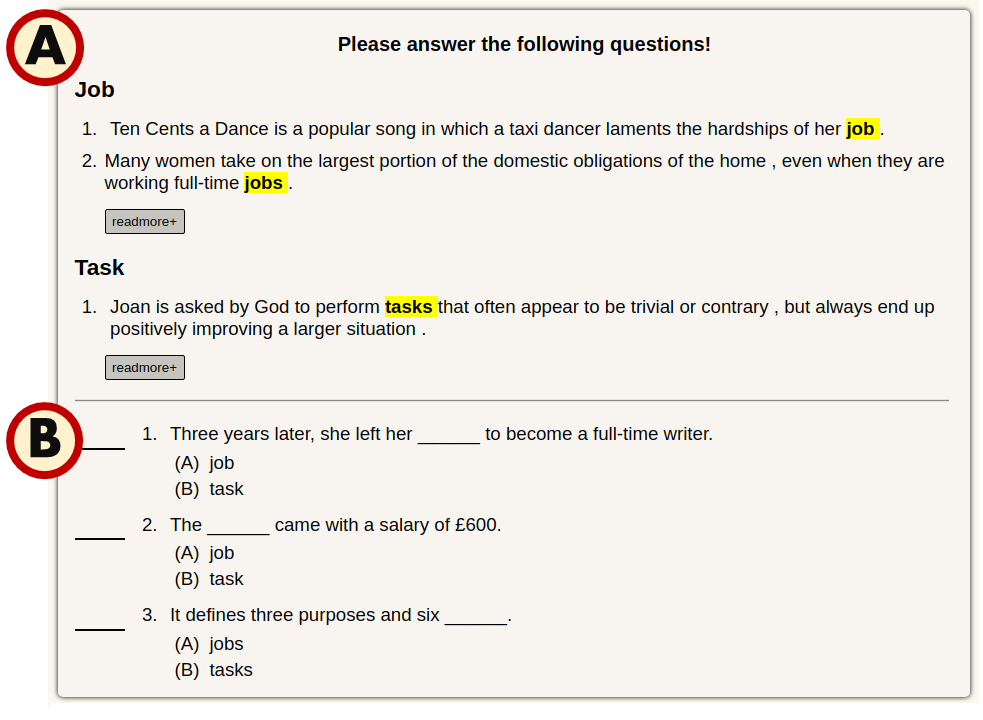}
    \caption{The interface of the user study contains two panel, (A) the example sentence panel and (B) the test panel. The example sentence panel will only be presented in the post-test.}
    \vspace{-5mm}
    \label{fig:user-study}
\end{figure}

We followed \citet{huang2017towards}'s learner study design with some modification.
The whole test consisted of a pre-test and a post-test section in a total of 80 minutes. 
The fill-in-the-blank multiple-choice question was used in both tests to examine students' understanding of near-synonym.
A total of 30 word pairs were used to create 30 question sets where each set contained three questions.
The questions are manually selected by an ESL expert from the wiki, Cambridge, or BBC dictionary.
Figure~\ref{fig:user-study} shows the interface of the post-test.
In the pre-test, only the test panel, as shown in Figure~\ref{fig:user-study} (B), was presented to students.
The students were asked to finish the randomly assigned 15 question sets in the pre-test and a background questionnaire.
During the post-test section, example sentences generated by \entshort, \fitbshort, or the random baseline will be presented in the example panel as shown in Figure~\ref{fig:user-study} A.
A maximum of three example sentences for each word can be obtained by clicking the \textit{readme} button.
The readme button can help us track how many example sentences were used for learning.
Note that the students were asked to answer the same question sets in the post-test so we can measure the improvement they made between the pre-test and the post-test. 
For each question set, the model used for sentence selection was also randomly assigned in order to prevent learners from getting tired from the useless example sentences.
Different from the sentence selection in Section~\ref{subsec:sentence-selection}, 
where all the combinations with the highest score in the quiz are selected, 
we picked the most common three example sentences from the combination to fulfill the experimental design. 
Here, we assume the most common three sentences for each word would be the best candidate in all the combinations. 

\begin{figure}[t]
    \centering
    \includegraphics[width=1.0\columnwidth]{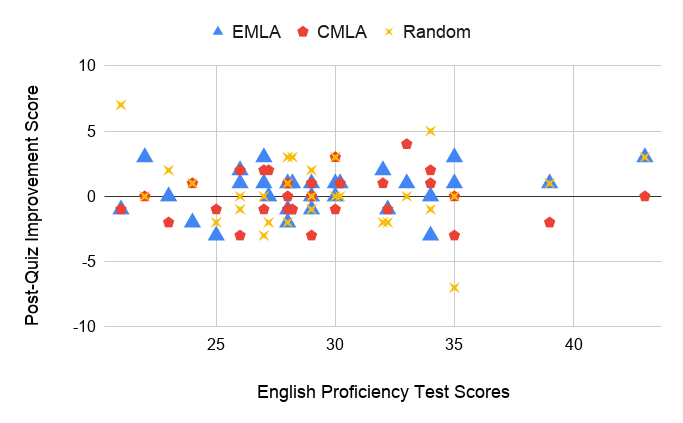}
    \caption{Improvement of 29 learner scores in respect to entailment modeling, context modeling, and random baseline. A total of 16 learners improved when learning on the material generated by entailment modeling.}
    \label{fig:user_result_general}
    \vspace{-0.2cm}
\end{figure}

\begin{table}[t]
    \centering \small
    \addtolength{\tabcolsep}{-0.07cm}
    \begin{tabular}{cc|c|c|c}
    \toprule \hline
                                \multicolumn{2}{c|}{}           & \textbf{\entshort} & \textbf{\fitbshort}  & \textbf{Random} \\ \hline
        \multirow{2}{*}{\textbf{Improvement}} &     Above & 0.75 & 0.42  & 0.00   \\ \cline{2-5}
                                     &     Below & 0.18 & -0.24 & 0.47   \\ \hline
                                     \multicolumn{5}{c}{}\\ \hline
        \multirow{2}{*}{\textbf{\# Examples}} &     Above & 4.34* & 4.43*  & 3.46*   \\ \cline{2-5}
                                     &     Below & 5.42 & 5.41 & 5.41   \\ \hline         
                                     \multicolumn{5}{c}{}\\ \hline
        \multirow{2}{*}{\textbf{Difficulty Rating}} &     Above & 2.40 & 2.36*  & 2.39   \\ \cline{2-5}
                                     &     Below & 2.58 & 2.68 & 2.47   \\ \hline \bottomrule
    \end{tabular}
    \addtolength{\tabcolsep}{0.07cm}
    \caption{Analysis of two groups. Above and Below stand for the above-average group and the below-average group respectively. \entshort helps the above-average group the most. We also find that the above-average group reads significantly fewer sentences than the below-average group. However, the below-average group rates the example sentences easier (scores range from 1 to 4 while 1 being ``too difficult'').}
    \vspace{-0.3cm}
    \label{tab:user-study-result}
\end{table}

\subsection{Results and Discussion}

When learning from example sentences from \entshort, 16 students improved.
Only 12 and 11 students improved when learning from \fitbshort and random baseline, suggesting that \entshort helped more.
Figure~\ref{fig:user_result_general} shows the students' improvement score versus proficiency score.


To further understand students' behaviors, we separated students into two groups using their English proficiency test scores.
Students whose test scores were lower than the average score were grouped into the below-average group and were considered having lower English proficiency, and vice versa.
The above-average group and the below-average group had 12 and 17 students respectively.
The average improvement scores of the two groups are shown in Table~\ref{tab:user-study-result}.
We can see the above-average students benefit more from example sentences while below-average benefit less or even confused by the example sentences. 
Again, \entshort helps above-average students the most.
The random baseline provides a mixed result, and even the above-average students got affected.
This echos results from \citet{huang2017towards} where students can still learn from the random example sentences but more effort is needed to fully understand the near-synonym and the outcome is unstable. 
In Figure~\ref{fig:user_result_general}, we can find that there are two outliers in the random baseline.
The one improved a lot is from the below-average group, and the other one worsen a lot is from the above-average group.
This evidence shows the uncertainty of the random baseline.


We investigated the learner's behavior during the post-test and their questionnaire response toward example difficulty. 
The result is also shown in Table~\ref{tab:user-study-result}.
The above-average students read significantly fewer examples while they also rate examples more difficult.
On the other hand, most of the below-average students read all the six examples and rate them relatively easier.
Though many above-average students improved in the post-test, we found that there are two of them read less than three examples and thus performed worse in the post-test.
Such a case suggests that reading a fair amount of example sentences is required to fully understand the near-synonym.

\vspace{-1mm}
\section{Conclusion}
\vspace{-1mm}
We introduce the learner-like agent, in particular \entshort, which differentiates the
helpfulness of learning materials using inference. 
Entailment modeling, unlike common context-based near-synonymous word
disambiguation, makes inferences to learn the relationship between the
example sentences and the question, similar to human behavior. 
Context modeling in the learner-like agent relies upon additional perturbed examples to mimic human
behavior, whereas \entshort already has this ability. The agent can be used to
evaluate the helpfulness of learning materials, or---more interestingly---to
select the best materials from a large candidate pool. We select good example
sentences in practice, which confirms the usefulness of modeling learner
behavior. Using the \entshort learner-like agent, we find more helpful learning
material for learners, as demonstrated by the learner study. These demonstrate the
usefulness of modeling learner behavior using an inference approach.
In the future, we would like to explore if the learner-like agent 
can be extended to materials and data beyond the example sentences for near-synonyms.

\section*{Acknowledgments}

This research was partially supported by the Ministry of Science and Technology of Taiwan under contracts MOST 108-2221-E-001-012-MY3 and MOST 109-2221-E-001-015-.

\bibliography{emnlp2020}
\bibliographystyle{acl_natbib}

\end{document}